# On Comparison between Evolutionary Programming Network based Learning and Novel Evolution Strategy Algorithm-based Learning

M. A. Khayer Azad, Md. Shafiqul Islam and M. M. A. Hashem[1]
Department of Computer Science and Engineering
Khulna University of Engineering and Technology
Khulna 9203, Bangladesh

## ABSTRACT

This paper presents two different evolutionary systems - Evolutionary Programming Network (EPNet) [1] and Novel Evolutions Strategy (NES) Algorithm [2]. EPNet does both training and architecture evolution simultaneously, whereas NES does a fixed network and only trains the network. Five mutation operators proposed in EPNet to reflect the emphasis on evolving ANN's behaviors. Close behavioral links between parents and their offspring are maintained by various mutations, such as partial training and node splitting. On the other hand, NES uses two new genetic operators - subpopulation-based max-mean arithmetical crossover and time-variant mutation. The above-mentioned two algorithms have been tested on a number of benchmark problems, such as the medical diagnosis problems (breast cancer, diabetes, and heart disease). The results and the comparison between them are also present in this paper.

## 1. INTRODUCTION

Now-a-days, neural network is used in. many applications such as robotics, missile projection, and rocket even in cooker. A number of algorithms are developed to train the network and evolve the architecture. Most applications use feed forward ANNS and the back-propagation (BP) training algorithm. The problem of designing a near optimal ANN architecture for an application remains unsolved.

There have been many attempts in designing ANN architectures automatically, such as various constructive (starts with minimal and adds) and pruning algorithms [3, 4] (opposite of constructive).

This paper describes a new evolutionary system, i.e., EPNet, for evolving feed forward ANNS. It combines. The architectural evolution with the weight learning. It also describcs a Novel Evolution Suategy (NES) Algorithm [4, 5, 6,12] that eliminates the problems which are encountered in contemporary ESs. NES uses a unique recombination operator which has a *collective interaction* of individuals within a particular subpopulation of the population. This algorithm also uses a time-variant Gaussian mutation scheme based on observed natural phenomena.

This paper also describes the comparison between EPNet and NES using the three medical diagnosis problems. In this paper it is shown that NES is comparatively better than EPNet.

## 2. EVOLVING ARTIFICIAL NEURAL NETWORK ARCHITECTURES

There are two major approaches to evolving ANN architectures. One is the evolution of "pure" architectures (i.e. architectures without weights). Connection weights will be trained after a near optimal architecture has been found. The other is the simultaneous evolution of both architectures and weights. It is clear that the evolution of pure architectures has difficulties in evaluating fitness accurately. As a result, the evolution would be inefficient.

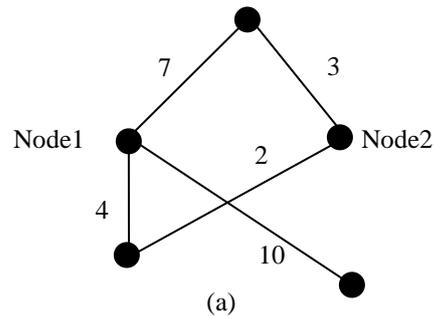

(a)

0100  1010  0010  0000  0111  0011

(b)

Fig.1: (a) An ANN and (b) Its genotypic representation, assuming that each weight is represented by 4 binary bits, Zero weight implies no connection.

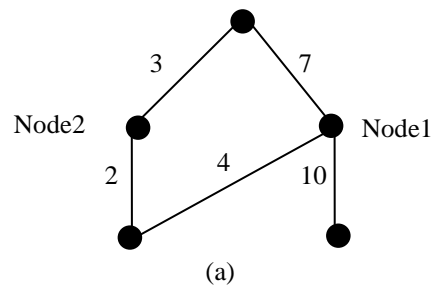

(a)

0010  0000  0100  1010  0011  0111

(b)

Fig.2: (a) An ANN which is equivalent to that given in figure 1(a) and (b) Its genotypic representation.

---
[1] Corresponding author E-mail: hashem@cse.kuet.ac.bd

This problem not only makes the evolution inefficient, but also makes crossover operators more difficult to produce highly fit offspring. It is unclear what building blocks actually are in this situation. For example, ANNS shown in Fig. 1(a) and Fig 2 (a) are equivalent, but they have different genotypic representations as shown by Fig. 1(b) and Fig. 2(b) using a direct encoding scheme.

### 3. EVOLUTIONARY PROGRAMMING NETWORK

Evolutionary Programming (EP's) emphasis on the behavioral link between parents and their offspring also matched well with the emphasis on evolving ANN behaviors, not just circuitry. In its current implementation, EPNet is used to evolve feed forward ANNs with sigmoid functions. However, this is not an inherent constrain. In fact, EPNet has minimal constraint on the type of ANNs which may be evolved. The major steps of EPNet can be described by Fig. 3, which are explained below [7, 8, 9]:

1. Generate an initial population of *M* networks at random. The number of hidden nodes and the initial connection density for each network are uniformly generated at random within certain ranges. The random initial weights are uniformly distributed inside a small range.
2. Partially train each network in the population on the training set for a certain number of epochs using a modified BP (MBP) with adaptive learning rates. The number of epochs, $K_0$, is specified by the user. The error value E of each network on the validation set is checked after partial training. If *E* has not been significantly reduced, then the assumption is that the network is trapped in a local minimum and the network is marked with "failure". Otherwise the network is marked with" success".
3. Rank the networks in the population according to their error values, from the best to the worst.
4. If the best network found is acceptable or the maximum number of generation has been reached, stop the evolutionary process and go to step 10. Otherwise continue.
5. Use the rank-based selection to choose one parent network from the population. If it mark is "success", go to step 6, or else go to step 7.
6. Partially train the parent network for $K_1$ epochs using the MBP to obtain an offspring network and mark it in the same way as in step 2, where $K_1$ is a user specified parameter. Replace the parent networks with the offspring in the current population and go to step 3.
7. First decide the number of hidden nodes $N_{hidden}$ to be deleted by generating a uniformly distributed random number between 1 and a user-specified maximum number. $N_{hidden}$ is normally very small in the experiments, no more than 3 in most cases. Then delete $N_{hidden}$ hidden nodes from the parent network uniformly at random. Partially train the pruned network by the MBP to obtain an offspring network. If the offspring network is better than the worst network in the current population, replace the worst by the offspring and go to step 3. Otherwise discard this offspring and go to step 8.

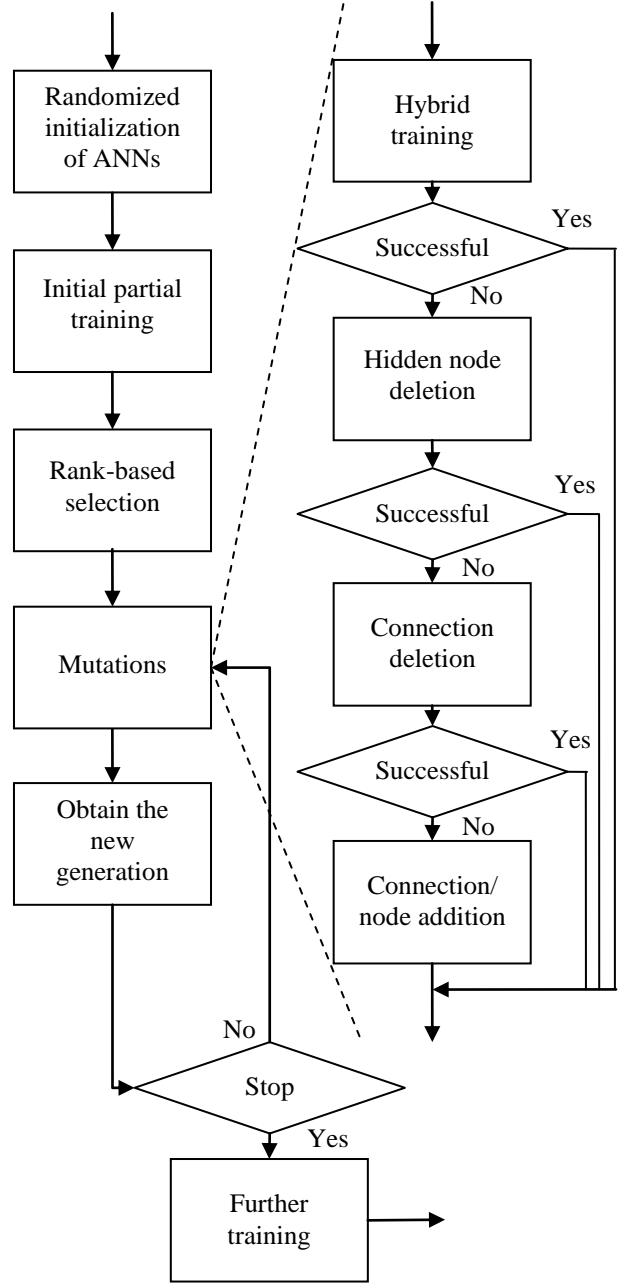

Fig. 3: Major steps of EPNet

8. Calculate the approximate importance of each connection in the parent network using the non-convergent method. Decide the number of connections to be deleted in the same way as that described in step 8. Randomly delete the connections from the parent network according to the calculated importance. Partially train the pruned network by the MBP to obtain an offspring network. If the offspring network is better than the worst network in the current population, replace the worst by the offspring and go to step 3. Otherwise discard this offspring and go to step 9.
9. Decide the number of connections and nodes to be added in the same way as that described in the step 7. Calculate the approximate importance of each virtual connection with zero weight. Randomly add the connections to the parent network to obtain Offspring 1 according to their importance. Addition of each node is implemented by splitting a randomly selected hidden node in the parent network. The new grown network after adding all nodes is offspring 2. Partially train offspring 1 and offspring 2 by the MBP to obtain a survival offspring. Replace the worst network in the current population by the offspring and go to step 3.
10. After the evolutionary process, train the best network further on the combined training and validation set until it "converges".

The above evolutionary process appears to be rather complex, but its essence is an EP algorithm with five mutations: hybrid training, node deletion, connection deletion, connection addition and node addition.

### 3.1 Encoding Scheme for Feedforward Artificial Neural Networks

The feed forward ANNs considered by EPNet are generalized multilayer perceptrons [10] (p.272-273). The architecture of such networks is shown in Fig. 4, where X and Y are inputs and outputs respectively.

$$x_i = X_i, 1 \leq i \leq m$$

$$net_i = \sum_{j=1}^{i-1} w_{kj} x_j, m < i \leq m + N + n$$

$$x_j = f(net_j), m \leq j \leq m + n + N$$

$$Yi = xi + m + N, 1 \leq i \leq n$$

where $f$ is the following sigmoid function:

$$f(z) = \frac{1}{1 + e^{-z}}$$

$m$ and $n$ are the number of inputs and outputs respectively, $N$ is the number of hidden nodes.

The direct encoding scheme is used in EPNet to represent ANN architectures and connection weights (including biases). EPNet evolves ANN architectures and weights simultaneously and needs information about every connection in an ANN. Two equal size matrices and one vector are used to specify an ANN in EPNet. The size of the two matrices is $(m + N + n) \times (m + N + n)$, where $m$ and $n$ are the number of input and output nodes respectively, and $N$ is the maximum number of hidden nodes allowable in the ANN. One matrix is the connectivity matrix whose entries can only be 0 or 1. The other is the corresponding weight matrix whose entries are real numbers.

### 3.2 Fitness Evaluation and Selection Mechanism

The fitness of each individual in EPNet is solely determined by the inverse of an error value defined by Eq. (1) [14] over a validation set containing $T$ patterns:

$$E = 100 \cdot \frac{o_{max} - o_{min}}{T.n} \sum_{t=1}^{T} \sum_{i=1}^{n} (di(t) - yi(t))^2 \quad (1)$$

where $o_{max}$ and $o_{min}$ are the maximum and minimum values of output coefficients in the problem representation, $n$ is the number of output nodes, and are desired and actual outputs of node $i$ for pattern $t$.

Eq.(1) was suggested by Prechelt [11] to make the error measure less dependent on the size of the validation set and the number of output nodes. Hence a mean squared error percentage was adopted. $o_{max}$ and $o_{min}$ were the maximum and minimum values of outputs [11].

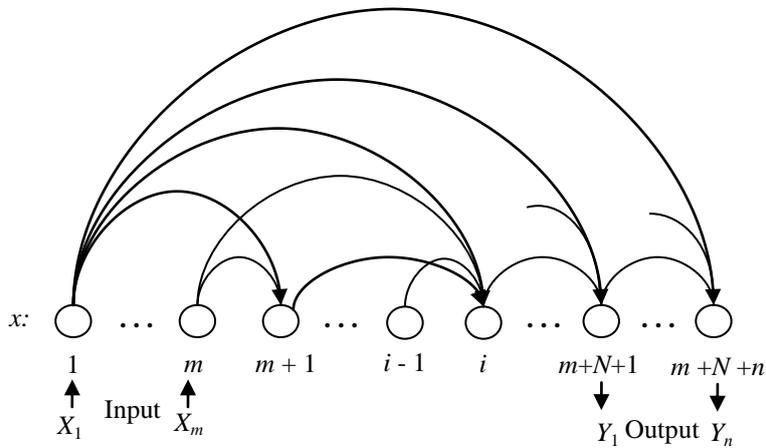

Fig. 4: A fully-connected feedforward artificial neural network [10] (pp.273)

### 3.3 Architecture Mutations

In EPNet, only when the training fails to reduce the error of an ANN will architectural mutations takes place. For architectural mutations, node or connection deletions are always attempted before connection or node additions in order to encourage the evaluation of small ANNs. Connection or node additions will be tried only after node or connection deletions fail to produce a good offspring. Using the order of mutations to encourage parsimony of evolved ANNs represents a dramatically different approach from using a complexity term in the fitness function. It avoids the time-consuming trial-and-error process of selecting a suitable coefficient for the regulation term.

#### 3.3.1 Hidden Node Deletion

Certain hidden nodes are first deleted uniformly at random from a parent ANN. The maximum number of hidden nodes that can be deleted is set by a user specified parameter. Then the mutated ANN is partially trained by the MBP. This extra training process can reduce the sudden behavioral change caused by the node deletion. If this trained ANN is better than the worst ANN in the population, the worst ANN will be replaced by trained one and no further mutation will take place. Otherwise connection deletion will be attempted.

#### 3.3.2 Connection Deletion

Certain connections are selected probabilistically for deletion according their importance. The maximum number of connections that can be deleted is set by a user-specified parameter.

Similar to the case of node deletion, the ANN will be partially trained by the MBP after certain connections have been deleted from it. If the trained ANN is better than the worst ANN in the population, the worst ANN will be replace by the trained one and no further mutation will take place. Otherwise node/connection addition will be attempted.

#### 3.3.3 Connection and Node Addition

Certain connections (with zero weights) are added to a parent network initialized with small random weights. The new ANN will be partially trained by the MBP and denoted as Offspring 1. The new ANN produced by node addition is denoted as Offspring 2 after it is generated, it will also be partially trained by the MBP. Then it has to compete with Offspring 1 for survival. The survived one will replace the worst ANN in the population.

### 4. NOVEL EVOLUTION STRATEGY

Two important variation (genetic) operators are based on some natural evidence of evolution for the NES algorithm.
1. Subpopulation-Based Max-mean Arithmetical Crossover (SB MAC).
2. Time-Variant Mutation (TVM).

### 4.1 Subpopulation-Based Max-mean Arithmetical Crossover

The parent population $\Pi(t)$ consisting of $\mu$ individuals is divided into $l$ subpopulations in each generation $t$ such that each subpopulation will have $\frac{\mu}{l}$ individuals. The individuals $\Psi_{j,\max}^t$ is defined as an elite individual that maximized a cost function, $f^{(t-1)}$ within the $j$-th subpopulation, and mean-individual $\overline{\Psi}_j^t$ (virtual parent) is created from the $j$-th subpopulation excluding the $\Psi_{j,\max}^t$.

Now the crossover operation is defined to produce two offspring $(\zeta_1^t, \zeta_2^t)$ as

$$\zeta_1^t = \alpha \Psi_{j,\max}^t + (1-\alpha)\overline{\Psi}_j^t \qquad (2)$$

$$\zeta_2^t = (1-\alpha)\Psi_{j,\max}^t + \alpha\overline{\Psi}_j^t \qquad (3)$$

where $\alpha$ is selected from URN[0,1] and $\alpha$ is sampled a new for each object variable of the individuals. The parameter $l$ is called an *exogenous parameter* of the method.

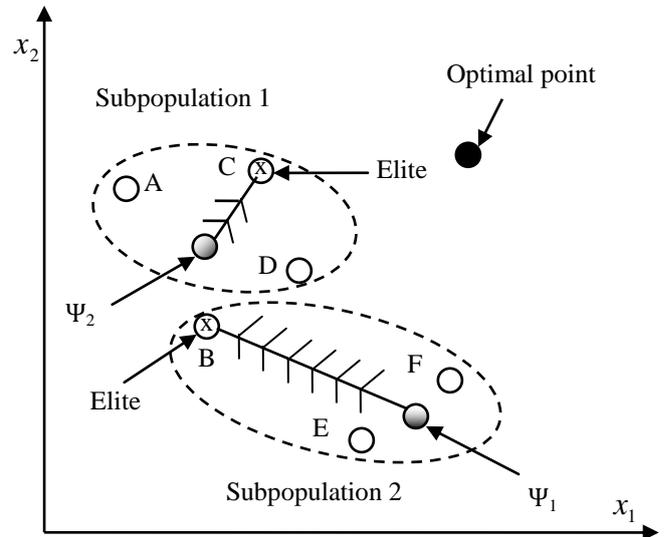

Fig. 5: Subpopulation–Based Max-mean Arithmetical Crossover (SBMAC).

### 4.2 Time-Variant Mutation

The TVM is defined for a child as that of ESs do $\zeta_j^t$ as

$$\xi_i' = \xi_i + \sigma(t).N_i(0,1) \qquad \forall i \in \{1,...,n\} \qquad (4)$$

where $N_i(.,.)$ indicates that the Gaussian random value with zero-mean and unity variance, and it is sample anew for each value of the index $i$. And $\sigma(t)$ is the time-variant mutation step generating function at the generation $t$, which is defined by

$$\sigma(t) = \left[1 - r^{(1-\frac{t}{T})^\gamma}\right] \qquad (5)$$

where $r$ is selected from URN[0,1], $T$ is the maximal generation number, $\gamma$ is a real-valued parameter determining the degree of dependency on the generations. The parameter $\gamma$ is also called an *exogenous parameter* of the method.

*4.3 NES Algorithm*

The general pseudo-code type structure of the proposed novel evolution strategy (NES) algorithm which utilizes the above mentioned two variation operators is shown in Fig. 6.

```
Algorithm_NES()
{
    t = 0;         /* Initialize the generation counter*/
    Initialize_Population();
    Evaluate_Population();
    while (NOT termination condition satisfied) do
        {
          Apply_SBMAC();/*Crossover operation*/
          Apply_TVM(); /*Mutation operation*/
          Evaluate_Population();
          Alternate_Generation();
          t + +;  /* Increase the generation counter*/
        }
}
```

Fig. 6: A pseudo-code structure of NES.

*4.3.1 Initial population*

The initial population, $\Pi(0)$ consisting of $\mu$ individuals, is generated by using a Uniform Random Number (URN) within desired domain of the object variables. After evaluating the $\mu$ individuals to their fitness function $\phi$, this population is considered as parents for the next generation.

*4.3.2 Crossover*

In the crossover the SBMAC is used to produce the offspring population. For each subpopulation, $\frac{\mu}{l}$ offspring are generated. Thus, $\mu$ numbers of offspring are generated for the $l$ subpopulations at the generation $t$.

*4.3.3 Mutation*

In the mutation phase, the TVM operator is used to mutate all variables of an offspring. Thus the offspring population undergoes this mutation scheme. It is ought to be taken care that initially this type of mutation might violate the domain of the object variables. In case of domain violations for any offspring, that offspring is left without mutation.

*4.3.4 Evaluation*

After mutation operation, each offspring $\zeta^{t}$ is evaluated its cost function (function) $\phi^{t}$ for a possible solution in each generation.

*4.3.5 Alternation of Generation*

In the alternation of generation, $(\mu + \mu)$-ES is used. That is, among $\mu^{t-1}$ parents which were evaluated at the former generation, and $\mu^{t}$ children which are evaluated in the current generation $t$, the $\mu^{t-1} + \mu^{t}$ individuals are ordered according to their cost function values and the best $\mu$ individuals will be selected for the next generation.

**5. EXPERIMENTAL STUDIES**

In order to evaluate EPNet's ability in evolving ANNs that generalize well, EPNet was applied to four real-world problems in the medical domain (i.e., breast cancer, diabetes, heart disease problem). All the data sets were obtained from the UCI machine learning benchmark repository (http://ics.uci.edu in directory/pub/machine-learning-database). The data sets are also applied in Novel Evolutionary Strategy (NES) algorithm.

*5.1 The Medical Diagnosis Problems*

The medical diagnosis problems mentioned above have the following common characteristics [11]:
- The input attributes are similar to a human expert would use in order to solve the same problem.
- The outputs represent either the classification of a number of understandable classes or the prediction of a set of understandable quantities.
- In practice, all these problems are solved by human experts.
- Examples are expensive to get. This has the consequence that the training sets are not very large.

*5.1.1 The Breast Cancer Data Set*

The breast cancer data set was originally obtained from W. H. Wolberg at the University of Wisconsin Hospitals, Madison. The data set contains 9 attributes and 699 examples of which 458 are benign examples and 241 are malignant examples.

*5.1.2 The Diabetes Data Set*

This data set was originally donated by Vincent Sigillito from Johns Hopkins University and was constructed by constrained selection from a larger database held by the National Institute of Diabetes and Digestive and Kidney Diseases. All patients represented in this data set are females of at least 21 years old and of Pima Indian heritage living near Phoenix, Arizona, USA. This is a two class problem with class value 1 interpreted as "tested positive for diabetes". There are a total of 768 examples are used of which 500 examples of class 1 and 268 of class 2. There are 8 attributes for each example.

*5.1.3 The Heart Disease Data Set*

This data set comes from the Cleveland Clinic Foundation and was supplied by Robert Detrano of the V. A. Medical Center, Long Beach, CA. The purpose of the data set is to predict the presence or absence of heart disease. This database contains 13 attributes, which have been extracted from a larger set of 75. There are 35 attributes for each example.

*5.2 Experimental setup*

All the data sets used by EPNet were partitioned into three sets: a training set, a validation set, and a testing set. In the following experiments, each data set was partitioned as follows:
- For the breast cancer data set, the first 349 examples were used for the training set, the following 175

examples for validation set, and the following 175 examples for testing set.
- For the diabetes data set, the first 384 examples were used for the training set, the following 192 examples for validation set, and the following 192 examples for testing set.
- For the heart disease data set, the first 134 examples were used for the training set, the following 68 examples for validation set, and the following 68 examples for testing set.

The input attributes of the diabetes data set and heart disease data set were rescaled to between 0.0 and 1.0 by a linear function. The output attributes of all the problems were encoded using a 1-of-*m* output representation for *m* classes.

### 5.3 Control Parameters

Control parameters in EPNet used in the experiments were: population size (20), initial connection density (1.0) (initial connection weights between -0.5 to +0.5), fixed bias (-1.5), learning rate (0.15), number of mutated hidden nodes (1), number of mutated connections (1 to 3). Number of hidden nodes of each individual in the initial population ranges: 1 to 3 (breast cancer); 2 to 8 (diabetes); 3 to 5 (heart disease). Number of epochs ($K_0$) for training each individual in the initial population was: 50 (breast cancer and diabetes) and 60 (heart disease). The number of epochs for the partial training during evolution (i.e., $K_1$) was 20 for the three problems. The number of epochs for training the best individual on the combined training and testing data set was 70 for all the problems. A run of EPNet was terminated if the average error of the population had not decreased by more than a threshold value $\varepsilon$ (0.01 for the three problems) after consecutive $G_0$ (10) generations or a maximum number of generations (500) were reached. These parameters were chosen after some limited preliminary experiments and not meant to be optimal.

The control parameters for NES were set to be the same for all the three problems: the population size (20), the initial connection density (1.0) (i.e., initial connection weights were between -0.5 to +0.5), initial bias (-0.5 to +0.5), the exogenous parameter, $\gamma$ (8.0), and the maximum number of generation, *T* (500), size of sub-group *l* (4). The numbers of hidden nodes for each individual in the population were: 3 (breast cancer), 8 (diabetes), and 5 (heart disease).

### 5.4 Experimental Results

All the results that obtained in the experiment are shown below using tables and figures; the error values are calculated using Eq. (1).

Table 1: Architectures of evolved Artificial Neural Networks by EPNet (over 30 runs)

| Data set | | Number of connections | Number of hidden nodes | Number of generations |
|---|---|---|---|---|
| Breast cancer | Mean | 19.6 | 2.4 | 105.3 |
| | SD | 4.6 | 0.53 | 30.1 |
| | Min | 12 | 2 | 60 |
| | Max | 28 | 4 | 170 |
| Diabetes | Mean | 33.6 | 3.9 | 135 |
| | SD | 9.3 | 1.2 | 39.3 |
| | Min | 21 | 2 | 60 |
| | Max | 52 | 7 | 220 |
| Heart disease | Mean | 120.6 | 4.3 | 253 |
| | SD | 39.4 | 1.5 | 68.8 |
| | Min | 48 | 2 | 130 |
| | Max | 198 | 7 | 290 |

Table 2: Accuracies of trained and evolved ANN by EPNet and only trained by NES (training)

| Data set | | EPNet (error) | NES (error) |
|---|---|---|---|
| Breast cancer | Mean | 0.018131 | 0.01324691 |
| | SD | 0.000144 | 0.00009674 |
| | Min | 0.01776 | 0.01321529 |
| | Max | 0.018857 | 0.01501733 |
| Diabetes | Mean | 0.016898 | 0.00683 |
| | SD | 0.000054 | 0.00001 |
| | Min | 0.016758 | 0.006827 |
| | Max | 0.016947 | 0.007107 |
| Heart disease | Mean | 0.018963 | 0.0151352 |
| | SD | 0.000226 | 0.0000256 |
| | Min | 0.01881 | 0.015133 |
| | Max | 0.020352 | 0.015678 |

Table 3: Validation of EPNet and NES

| Data Set | EPNet | NES |
|---|---|---|
| Diabetes | 64.1% | 64.6% |

Table 4: Testing of EPNet and NES (Wrong classification in percentage)

| Data Set | EPNet | NES |
|---|---|---|
| Diabetes | 24% | 25% |

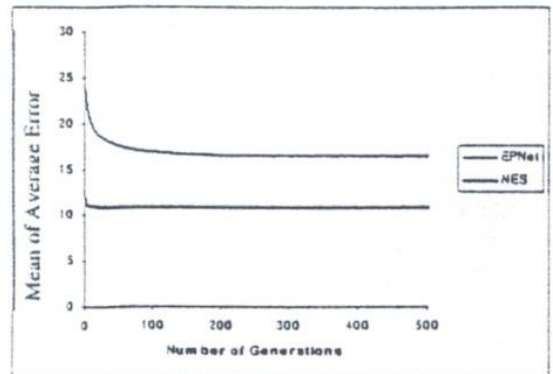

Fig. 7: Evolution histories of ANN by EPNet and NES for the Breast Cancer problem.

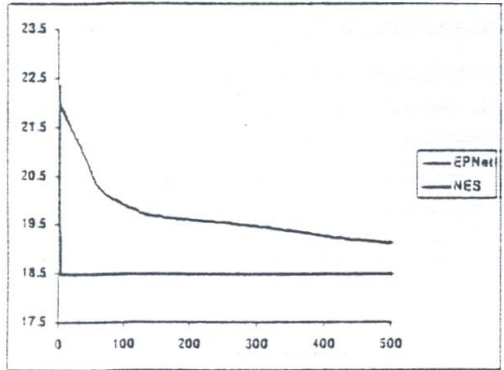

Fig. 8: Evolution histories of ANN by EPNet and NES for the diabetes problem.

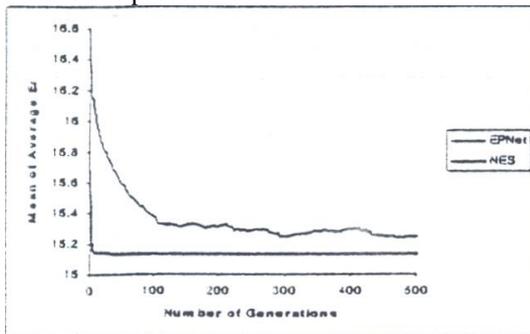

Fig. 9: Evolution histories of ANN by EPNet and NES for the Heart Disease problem.

### 6. CONCLUSIONS

In order to reduce the noise in fitness evaluation, EPNet evolves ANN architectures and weights simultaneously. On the other hand a novel evolution strategy (NES) learning algorithm trained the weights of a fixed architecture. The evolution simulated by EPNet is closer to the Lamarckian evolution than to the Darwinian one learned weights and architectures in one generation are inherited by the next generation. This is quite different from most genetic approaches where only architectures not weights are passed to the next generation.

EPNet encourages parsimony of evolved ANNs by ordering its mutations, rather than using a complexity (regularization) term in the fitness function. It avoids the tedious trial-and-error process to determine the coefficient for the complexity term. Novel evolution strategy (NES) algorithm also tested on a number of benchmark problems, including the three medical diagnosis problems. This algorithm utilized two new genetics operators – subpopulation-based max-mean arithmetical crossover and time-variant mutation – were abstracted based on some natural metaphor and biological observations, which are closely resembled to natural evolved systems. This algorithm came into view to be responded as simple as inexpensive.

From the experimental results we see that NES gives a fast decay than that of EPNet and NES algorithm that only trains a fixed network gives better accuracy than that of EPNet for the three medical diagnosis problems, for the training data set and validation data set but for the testing data set NES gives comparatively less accuracy than that of EPNet.